\documentclass[letterpaper]{article} % DO NOT CHANGE THIS
\usepackage{aaai2026}
\makeatletter

\def\copyright@text{}
\makeatother

\usepackage{times}  % DO NOT CHANGE THIS
\usepackage{helvet}  % DO NOT CHANGE THIS
\usepackage{courier}  % DO NOT CHANGE THIS
\usepackage[hyphens]{url}  % DO NOT CHANGE THIS
\usepackage{graphicx} % DO NOT CHANGE THIS
\usepackage{natbib}  % DO NOT CHANGE THIS AND DO NOT ADD ANY OPTIONS TO IT
\usepackage{caption} % DO NOT CHANGE THIS AND DO NOT ADD ANY OPTIONS TO IT
\usepackage{algorithm}
\usepackage{algorithmic}

\usepackage{newfloat}
\usepackage{listings}
\DeclareCaptionStyle{ruled}{labelfont=normalfont,labelsep=colon,strut=off} % DO NOT CHANGE THIS
\floatstyle{ruled}
\newfloat{listing}{tb}{lst}{}
\floatname{listing}{Listing}
\usepackage{tabularx}
\usepackage{amsmath}

\title{Attention Gathers, MLPs Compose: A Causal Analysis of an Action-Outcome Circuit in VideoViT}
\author {
    Sai V R Chereddy
}
\affiliations{
    Independent Researcher\\
    saivivaswanthreddy@alumni.usc.edu
}

\begin{document}

\maketitle

\begin{abstract}
The paper explores how video models trained for classification tasks represent nuanced, hidden semantic information that may not affect the final outcome, a key challenge for Trustworthy AI models. Through Explainable and Interpretable AI methods, specifically mechanistic interpretability techniques, the internal circuit responsible for representing the action's outcome is reverse-engineered in a pre-trained video vision transformer, revealing that the "Success vs Failure" signal is computed through a distinct amplification cascade. While there are low-level differences observed from layer 0, the abstract and semantic representation of the outcome is progressively amplified from layers 5 through 11. Causal analysis, primarily using activation patching supported by ablation results, reveals a clear division of labor: Attention Heads act as "evidence gatherers", providing necessary low-level information for partial signal recovery, while MLP Blocks function as robust "concept composers", each of which is the primary driver to generate the  "success" signal. This distributed and redundant circuit in the model's internals explains its resilience to simple ablations, demonstrating a core computational pattern for processing human-action outcomes. Crucially, the existence of this sophisticated circuit for representing complex outcomes, even within a model trained only for simple classification, highlights the potential for models to develop forms of 'hidden knowledge' beyond their explicit task, underscoring the need for mechanistic oversight for building genuinely Explainable and Trustworthy AI systems intended for deployment.
\end{abstract}

% \keywords{Video Vision Transformers \and Mechanistic Interpretability \and Causal Analysis \and Eliciting Latent Knowledge}

\section{1. Introduction}
Inspired by Vision Transformers (ViT) and their state-of-the-art performance in image classification tasks, \citet{arnab2021vivit} translated a pure transformer-based architecture for video classification tasks as well, introducing the Video Vision Transformer (ViViT) that focuses on the interplay of spatio-temporal tokens. However, even with their ever-increasing accuracy and capabilities, vanilla ViViT (and its variants) and other video models in general, fail to explain their reasoning for the output that they provide. This “black box” problem is persistent across modalities, including language and image but more so in videos, and lack of understanding of models’ internals and “hidden cognition” within its specified task environment, act as a critical barrier in humans deeming such deployed AI as trustworthy \citep{vonEschenbach2021transparency}.

Mechanistic Interpretability (or MechInterp) is a leading paradigm for addressing this opacity. MechInterp aims to completely specify a neural network’s computations ie; reverse-engineer the models' internals to gain a deeper understanding of the models’ thought-process \citep{bereska2024mechanistic}. Although there have been several techniques explored both in observational analysis (like feature visualization, logit lens, probes) and causal interventions (ablations, activation patching, hypothesis testing), majority of the work has been limited to language models and to some extent vision models. While \citet{olah2020zoom} lays foundations for interpretability in domains aside from text, rarely have these experiments been translated to the video frontier and this is primarily because moving frames brings about unique 3-dimensional spatio-temporal challenges making video interpretability a harder problem to solve \citep{anichenko2024interpretable}.

\par

This paper aims to address the concern of trustworthy video AI models in deployment, through mechanistic interpretability techniques applied to a 12-layer vanilla video vision transformer. The model is pre-trained on Kinetics-400 \citep{kay2017kinetics}, which is a basic human-action dataset, scrutinized in a simplistic experimental setup which involves using a contrastive pair of videos (the strike ball and gutter ball) which the model correctly classifies under the same class label of “bowling” even with different internal representations. Despite the identical and correct final output, mechanistic analysis reveals the model's deeper, hidden understanding of the contrasting interior mechanisms.

\par

The contributions of this paper are threefold.  First, an observational (attention visualization) and quantitative (delta analysis) evidence is provided revealing that a pre-trained Video Vision Transformer internally represents nuanced action outcomes (strike vs. miss) distinctly, even when its final classification labels for these outcomes are aligned with ground truth. Second, the paper demonstrates a methodology combining delta analysis on contrastive video pairs to locate the internal outcome signal and activation patching (measuring causal logit effects) to determine the functional roles of specific circuit components (Attention vs. MLP blocks). Finally, the setup reverse-engineers the core computational mechanism which distinguishes final action-outcomes, providing strong causal evidence that mid-layer MLP blocks act as the primary 'concept composers' driving this internal representation, refining the initial hypothesis of a simpler division of labor i.e. Attention Gathers, MLPs Compose.

\section{2. Related Work}

\subsection{2.1. Video Vision Transformers}
\label{sec:headings}

The success of the transformer architecture across modalities, firstly in text and then in vision (challenging the dominance of CNN) is fairly evident in today's world (\citep{vaswani2017attention}, \citep{dosovitskiy2020image}) . Translating this to moving frames, the Video Vision Transformers (ViViT) which is a pure transformer-based model for video classification is also seen to produce state-of-the-art perfomances, making it common for computer vision tasks across industries. Examples include autonomous driving perception, quality control in manufacturing, and automated medical image analysis. As these models are deployed in increasingly high-stakes domains, understanding their internal reasoning and ensuring their reliability becomes paramount for establishing trust \citep{meyers2023safety}. Structurally, the model extracts spatio-temporal tokens from the input video that are then encoded by a series of transformer layers. While \citet{arnab2021vivit} introduced several transformer models, this experiment focuses on spatio-temporal attention with tubelet embeddings that span across the “Joint Space-Time” dimensionalities. The architecture also encodes a special [CLS] token or classifier token that propagates throughout the model encoding information about the final output class at each layer. This provides a valuable target for interpretability techniques, allowing analysis of how the model's prediction evolves layer by layer.

\subsection{2.2. Mechanistic Interpretability}
\label{sec:headings}

Understanding inner workings of AI systems is critical for value alignment and safety, especially more so in deployment \citep{gabriel2020artificial}. A promising approach for this is Mechanistic Interpretability (or MechInterp), which is the reverse-engineering of computational mechanisms and representations learned by neural networks into human-understandable algorithms and concepts through a granular, causal analysis \citep{bereska2024mechanistic}.  This granular analysis involves understanding of features, neurons, layers, and connections, offering an intimate view of operational mechanics. The ideologies of MechInterp are broadly classified into two categories: Observational analysis and causal interventions. Observational techniques like logit lens, attention visualisations, structured probes and SAEs take a simple “look and understand” approach which are mainly to understand what parts of the model contribute to the final outcome and in what capacity. On the other hand, causal intervention studies deal directly with the model internals. Approaches like ablations and activation patching change the model internals to understand what differences did it make in the final outcome. This helps understand what the model is learning and which part of the system is responsible for perceived deceptions. Additionally, this causal approach offers a richer level of understanding than standard baselines like Saliency Maps, Integrated Gradients, or Concept Activation Vectors (CAVs). While such attribution methods primarily approximate which input features drive a prediction, mechanistic interpretability causally isolates "how" internal components transform those features. By intervening on specific hidden states, we can disentangle the functional roles of different architectural blocks, such as distinguishing between 'evidence gathering' (Attention) and 'concept composition' (MLPs) which offers a structural distinction often obscured in gradient-based attribution heatmaps.

\subsection{2.3. Eliciting Latent Knowledge}
\label{sec:headings}

A central challenge in ensuring the trustworthiness of deployed AI systems is understanding their internal states, especially when outputs might be misleading or insufficient. The field of Eliciting Latent Knowledge (ELK) specifically aims to address this by finding patterns in a model's activations that reliably track the truth, even when the model's explicit output cannot be trusted \citep{burns2022discovering}. This involves moving beyond surface behavior to potentially bypass deception or misalignment. Similarly, secret elicitation research focuses on uncovering knowledge a model possesses but does not verbalize. To study these phenomena in controlled settings, researchers like \citet{turner2025model} often create "model organisms", models intentionally trained to exhibit specific failure modes or hide information. For example, \citet{mallen2023eliciting} developed "quirky" LMs fine-tuned to make systematic errors in specific contexts (e.g., when the name "Bob" is present), using them to test if linear probes could still extract correct knowledge from internal activations. They found probes in middle layers often represented knowledge independently of the final output. \citet{cywinski2025eliciting} trained "secret-keeper" LLMs designed to possess and apply knowledge (like a secret word or user gender) while denying awareness when asked directly. They used these models to benchmark various black-box (e.g., prefill attacks ) and white-box (e.g., logit lens, Sparse Autoencoders ) elicitation techniques.

\section{3. Methodologies}
\subsection{3.1. Model \& Dataset: }
\label{sec:headings}

\subsubsection{Model:} The specific model used is the “google/vivit-b-16x2-kinetics400” which is open-sourced pre-trained on Kinetics-400 human-action dataset, available on HuggingFace. It is a base architecture of Video Vision Transformer (vivit-b) having 12 layers and attention heads similar to ViT-B. The 16x2 refers to tubelet embeddings with 16x16 spatial patch size and 2x temporal patch size, enabling the model to encode and process videos by dividing them into spatio-temporal “tubelets”.

\begin{figure*}[t]
  \centering
  \includegraphics[width=0.9\textwidth]{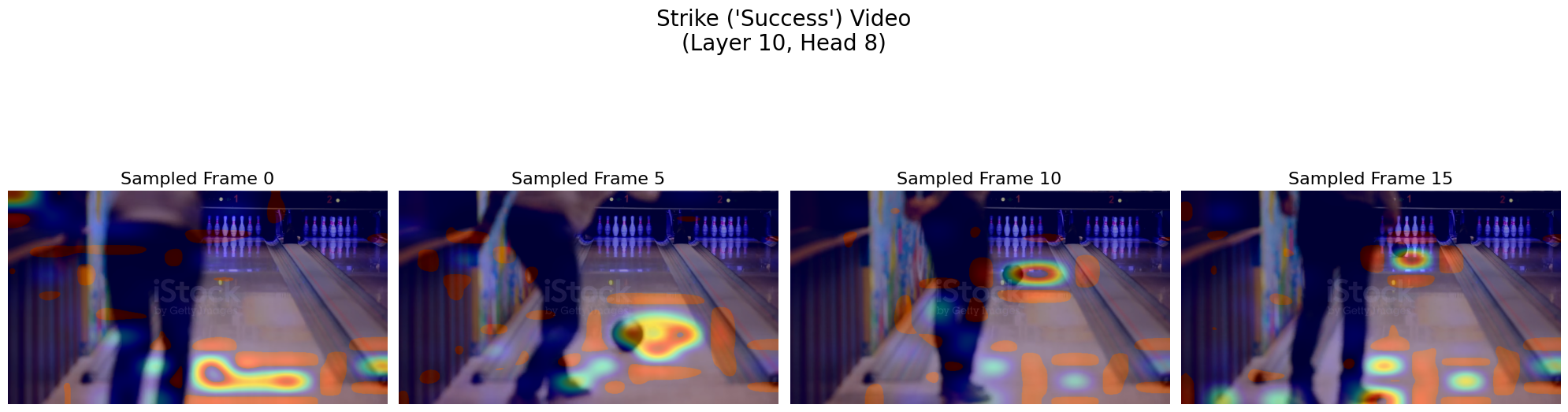}
  \includegraphics[width=0.9\textwidth]{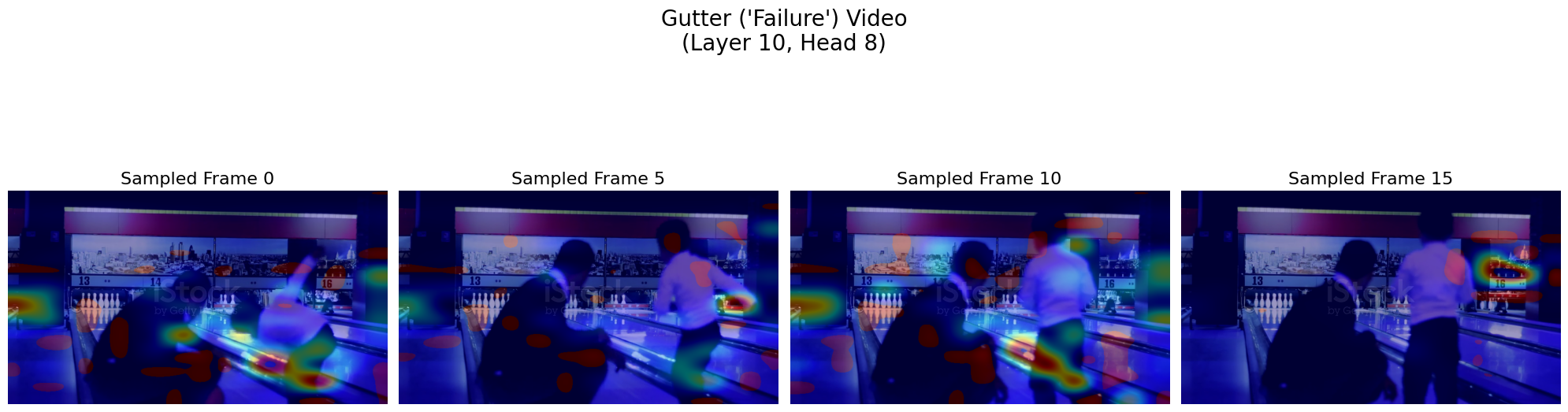}
  \caption{Qualitative visualization of CLS Token attention (Layer 10, Head 8) for the (a) 'Strike' and (b) 'Gutter' videos. This head functions as one of the semantic "outcome-detectors".}
  \label{fig:labels}
\end{figure*}

\subsubsection{Dataset:} 

The experiment utilizes a custom-generated minimal contrastive pair of 10-second videos from the Kinetics-400 \citep{sammani2023visualizing} "bowling" class (Label: 31): a positive "strike" run (ie; ball hits the pins) and a negative "gutter" run (ie; ball goes down the gutter). During preprocessing, frames are uniformly sampled with a stochastic temporal start index (random jitter) to align with standard training protocols. While this sampling typically introduces minor variance between runs, a fixed random seed (42) was used for all reported experiments to ensure exact reproducibility. The custom dataset and analysis code are available from the authors upon request.

\subsection{3.2. Observational Interpretability Analysis:}
\label{sec:headings}

This set of experiments pertained to understanding and merely observing the model internals to understand where the model focused when subjected to the input videos and what part of its architecture contributed to the final classification outcome \citep{joseph2023vit}.

\subsubsection{3.2.1. Direct Logit Attribution (DLA):} This is a technique for interpreting the output activations of model components. In language models, this is done in the vocabulary space but with ViViT’s architecture, we can focus directly on the [CLS] or the classifier token. By studying the logits in [CLS] token, DLA offers insights into which layer contributed most towards the final logits of the output class.

\subsubsection{3.2.2. Token-wise heatmap Visualization:} Focusing on the target class, this heatmap is generated by visualizing the contribution of each spatio-temporal token across all frames to the model's internal representation, offering insights into which parts of the input video the model attends to most strongly. 

\subsubsection{3.2.3. CLS Token Attention Visualization:} This heatmap (shown in Figure 1) gives us insights into the attention scores of the [CLS] token focusing on attention heads within a specific layer. The ViViT model divides each frame into patches making it easier to visualise the attention of each part of the image. Subsequently, an attention overlay was also plotted as seen in Figure 1 which gave insights into which part of the input video the model focuses on at a given frame.

\subsubsection{3.2.4. Linear Probe Analysis:} As an initial attempt to find a signal distinguishing the two outcomes, linear probes (a simple logistic regression classifier) were trained to distinguish between the activations of the [CLS] token from the "strike" run versus the "gutter" run \citep{zhao2024explainability}. This was done for each of the 12  layers of the VideoViT model with the goal to determine if the internal semantic concepts representing the outcome were linearly separable in the residual stream without more granular intervention.

\begin{figure*}[t]
    \centering
    \begin{minipage}[b]{0.45\linewidth}
        \centering
        \includegraphics[width=\linewidth]{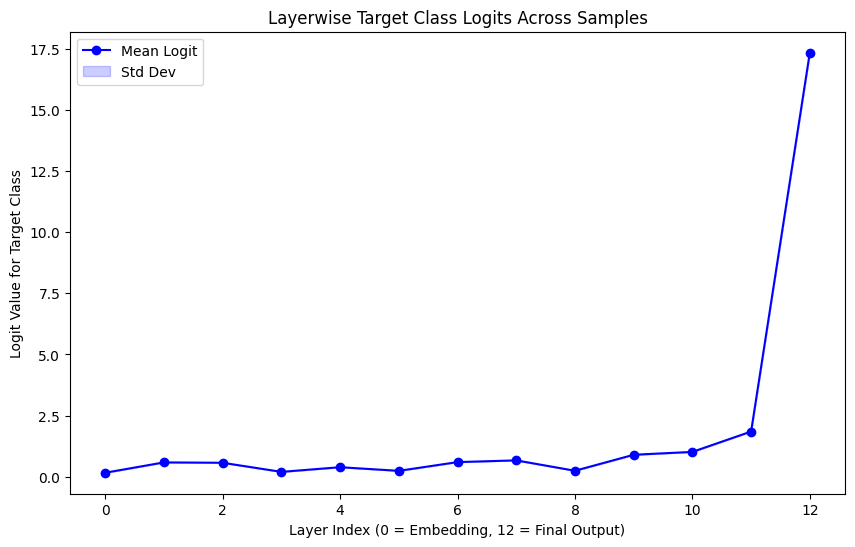}
    \end{minipage}
    \hfill
    \begin{minipage}[b]{0.45\linewidth}
        \centering
        \includegraphics[width=\linewidth]{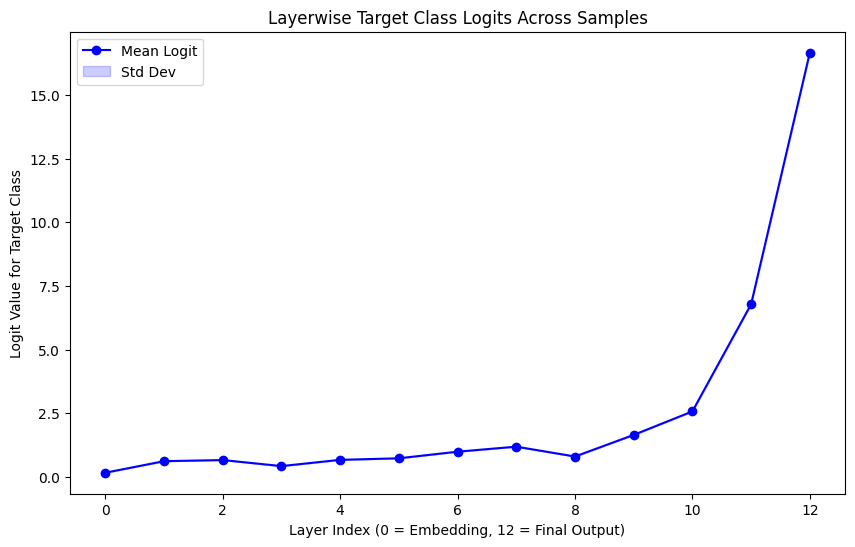}
    \end{minipage}
    \caption{Contribution of each layer towards the final logits of target class across frames for (a)strike and (b)gutter runs.}
    \label{fig:4}
\end{figure*}

\subsection{3.3. Signal Identification -  Delta Analysis:}While the initial observational analysis offered several insights into the model’s attention on the input frames and contributions of parts of model's architecture contributing towards the final classification output, they showed limited intuitions towards the semantic representation and hidden cognition in the models internals (detailed in sections 4.1 and 4.2). Hence, delta analysis was used to precisely locate the internal representation of the “Success vs Failure” signal calculated using the activation deltas like so: 

$\Delta = act_{strike} - act_{gutter}$

The L2 norm of this delta at each layer was used to  quantify the ”signal strength” and filter out noise; this calculation helped produce the signal amplification graph shown in Figure 5.

\subsection{3.4. Causal Analysis:}
This set of experiments included tinkering with the activations of the model at specific set of points to understand how much of these interventions would cause a change in the models final output i.e; to causally determine the function of specific model components (\citep{heimersheim2404use}, \citep{meloux2025everything}).

\subsection{3.4.1. Component Ablation:}
To rigorously test the causal necessity of salient visual features, an Automated Top-K Ablation strategy was used. The top $k\%$ of tokens (where $k=10$) were systematically zeroed out where the tokens that had the highest positive contribution to the target class logit across the entire spatio-temporal volume. This approach allows us to determine if the model's classification relies on these specific "action hotspots" or if it is robustly distributed across the global scene context.

\subsection{3.4.2. Activation Patching:} In this, activations were systematically patched (i.e; copied) from the “strike” run into the “gutter” run for each component (Attention vs MLP) at each layer. The goal was to measure the percentage of the final Layer 11 'Success vs. Failure' signal delta (identified via delta analysis) that was recovered by patching each single component.

\section{4. Results}

\subsection{4.1. Observational Interpretability Results}
\subsubsection{4.1.1. Direct Logit Attribution (DLA):} DLA helped in identifying which model layers contributed most to the final logit predicting the 'bowling' classification. As seen in Figure 2, from layer 9 and beyond the model becomes increasingly confident in its task to classify the input video.

\subsubsection{4.1.2. Token-wise heatmap Visualization:} As shown in figure 3, the tokens most directly contributing to the final output are spread across a number of frames, but fairly central within the picture which aligns well with the input video selected. The majority of these tokens correspond to the rolling ball and the ball-pin interaction in the “strike” video.

\begin{figure}[H]
  \centering
  \includegraphics[width=0.4\textwidth]{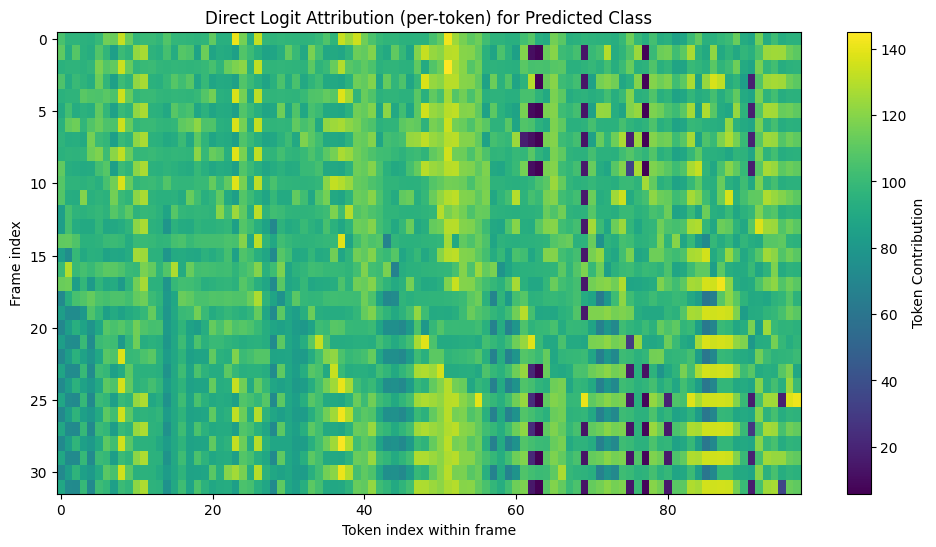}
  \caption{Token-wise heatmap across frames in "strike run" indicating contribution of each token to the final predicted output class}
  \label{fig:image8}
\end{figure}

\subsubsection{4.1.3. CLS Token Attention Visualization:} Similar to the overall token-wise heatmap seen in Figure 3 and attention overlay in Figure 1, the attention head 8 in layer 9 for the [CLS] token focuses on the ball-pin interaction for the “strike” run corresponding to patches 60-80 as seen in Figure 4.

\begin{figure}[H]
  \centering
  \includegraphics[width=0.45\textwidth]{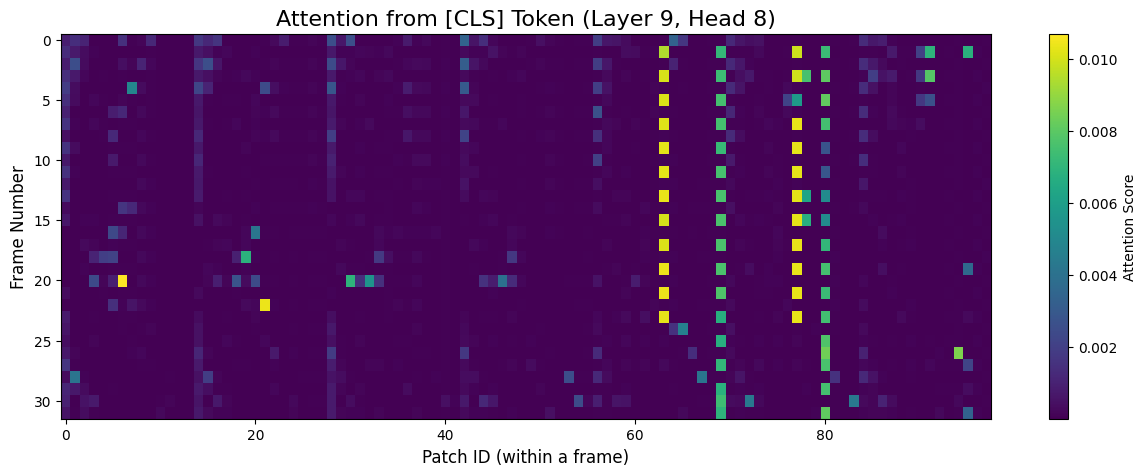}
  \caption{Attention visualisation heatmap for [CLS] token at Layer 9, Head 8.}
  \label{fig:heatmap}
\end{figure}

\begin{figure*}[t]
  \centering
  \includegraphics[width=0.65\textwidth]{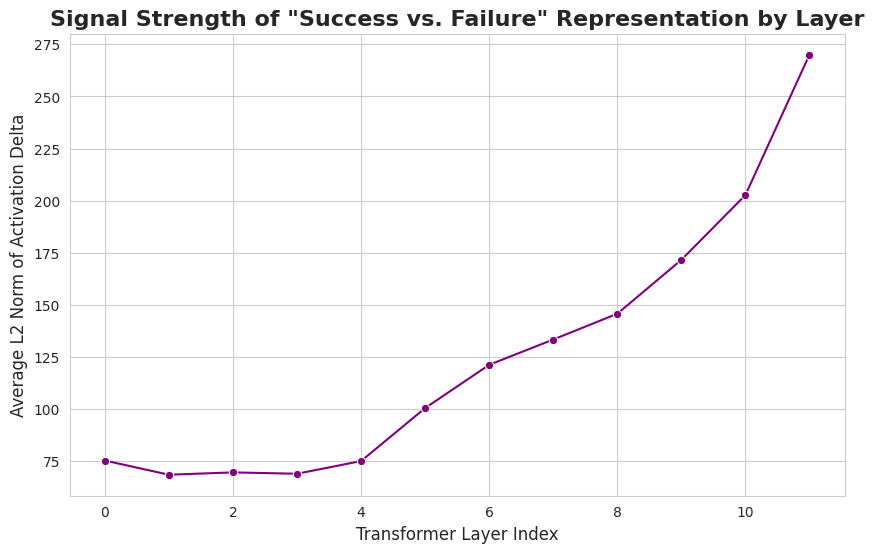}
  \caption{The average L2 norm of the activation delta between the "strike" and "gutter" runs. The sharp rise from layer 5 onwards shows the amplification cascade.}
  \label{fig:fig1}
\end{figure*}

To investigate this mechanism qualitatively, we visualized the attention from the specialized head of the next layer (Layer 10, Head 8) directly onto the video frames and this head acts as one of the semantic 'outcome-detectors' as seen in the attention overlay figure above.

In the 'Success' video (Figure 1a), the head's attention (red heatmap) dynamically tracks the ball's path (Frame 10) before snapping to the pins at the moment of impact (Frame 15).

Conversely, in the 'Failure' video (Figure 1b), the same head correctly identifies the failure mode: its attention focuses on the gutter (Frame 5) and subsequently settles on the untouched pins (Frame 15). This provides first visual evidence that the model might have learned a robust, high-level representation for both outcomes. That being said however, a numerical evidence was still lacking."

\begin{table*}[t]
\centering
\setlength{\tabcolsep}{4pt}
\renewcommand{\arraystretch}{1.1}

\begin{tabular*}{0.8\linewidth}{|c|c|c|c|}

\hline
\textbf{Prediction Rank} & \textbf{Original Prediction (Logit)} & \textbf{Ablated Prediction (Logit)} & \textbf{Change in Logits} \\
\hline
1 & Bowling (16.9954) & Bowling (16.6583) & -0.3371 \\
2 & Throwing Ball (9.0371) & Throwing Ball (9.1341) & 0.0970 \\
3 & Playing Cricket (6.2185) & Playing Cricket (6.4343) & 0.2158 \\
4 & High Kick (5.9146) & High Kick (5.7234) & -0.1912 \\
5 & Dancing Ballet (5.7508) & Dancing Ballet (5.7579) & 0.0072 \\
\hline
\end{tabular*}

\caption{Comparison of logits before and after ablation in "strike run".}
\label{tab:appendix_metrics}
\end{table*}

% ==================== Strike Video Analysis (Seed 42) ====================
% Rank  | Class Name                | Original   | Ablated    | Change    
% ---------------------------------------------------------------------------
% 1     | LABEL_31                  | 16.9954    | 16.6583    | -0.3371   
% 2     | LABEL_357                 | 9.0371     | 9.1341     | 0.0970    
% 3     | LABEL_227                 | 6.2185     | 6.4343     | 0.2158    
% 4     | LABEL_152                 | 5.9146     | 5.7234     | -0.1912   
% 5     | LABEL_84                  | 5.7508     | 5.7579     | 0.0072    

% ==================== Gutter Video Analysis (Seed 42) ====================
% Rank  | Class Name                | Original   | Ablated    | Change    
% ---------------------------------------------------------------------------
% 1     | LABEL_31                  | 16.5200    | 16.5044    | -0.0156   
% 2     | LABEL_357                 | 6.3758     | 6.6308     | 0.2551    
% 3     | LABEL_150                 | 4.8109     | 5.1653     | 0.3544    
% 4     | LABEL_227                 | 4.6217     | 4.6968     | 0.0751    
% 5     | LABEL_48                  | 4.2289     | 4.3045     | 0.0755    

\begin{table*}[t]
\centering
\setlength{\tabcolsep}{4pt}
\renewcommand{\arraystretch}{1.1}

\begin{tabular*}{0.8\linewidth}{|c|c|c|c|}
\hline
\textbf{Prediction Rank} & \textbf{Original Prediction (Logit)} & \textbf{Ablated Prediction (Logit)} & \textbf{Change in Logits} \\
\hline
1 & Bowling (16.5200) & Bowling (16.5044) & -0.0156 \\
2 & Throwing Ball (6.3758) & Throwing Ball (6.6308) & 0.2551 \\
3 & Headbutting (4.8109) & Headbutting (5.1653) & 0.3544 \\
4 & Playing Cricket (4.6217) & Playing Cricket (4.6968) & 0.0751 \\
5 & Throwing Baseball (4.2289) & Throwing Baseball (4.3045) & 0.0755 \\
\hline
\end{tabular*}

\caption{Comparison of logits before and after ablation in "gutter run".}
\label{tab:appendix_reload_metrics}
\end{table*}

\subsubsection{4.1.4. Linear Probe Analysis:} The logistic regression linear probe showed a misleading 100\% accuracy from Layer 0. This failed experiment indicated that the probe was acting as a 'fingerprint scanner' on superficial and spurious differences rather than identifying the underlying semantic concept.

\label{sec:headings}
\begin{figure}[H]
  \centering
  \includegraphics[width=0.43\textwidth]{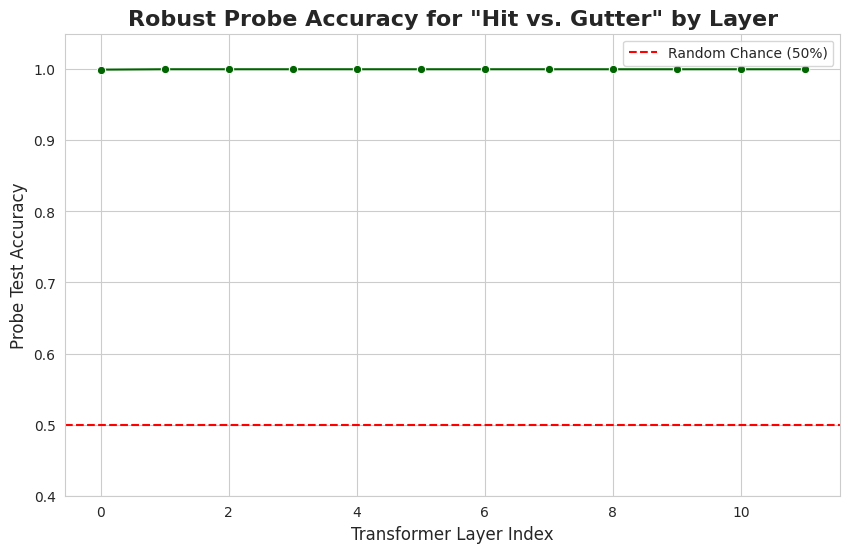}
  \caption{The layer-wise differentiating accuracy of the trained probe showed 100\% accuracy in distinguishing between the "strike" and "gutter" runs, acting as a superficial fingerprint scanner.}
  \label{fig:image3}
\end{figure}

\subsection{4.2. From Failed Experiments to a Clear Signal}

Initial explorations using Direct Logit Attribution (DLA) and CLS Token Visualization were successful in identifying which spatio-temporal region in the video the model considered important (i.e., the ball-pin interaction), but lacked insights into how this information was processed or represented semantically. To address this limitation, the L2 norm of the activation delta was analyzed:

\begin{equation}
\Delta = act\_strike -act\_gutter
\end{equation} 

between the two videos. As shown in Figure 5, this approach effectively filtered out low-level noise, revealing a clear "signal amplification cascade" from layer 5 through 11. Specifically, the average layer L2 norm shows a consistent increase of over 300\%, rising from approximately 75 at layer 5 to over 250 in the final layer, pinpointing the location of the potential outcome-computing circuit. 

Consequently this delayed emergence in delta analysis, in contrast to linear probe analysis (which detected superficial differences at Layer 0), signalled a circuit which is not tracking low-level spurious features (e.g., background texture), but is instead computing a higher-level semantic abstraction deep within the network.

\subsection{4.3. Component Ablations: Evidence for Circuit Separation}
Having established the existence of the outcome signal, it was tested to see whether the explicit classification ("Bowling") relied on the specific visual features associated with this action. An Automated Top-K Ablation strategy was utilized, using Direct Logit Attribution (DLA) to identify and zero out the top 10\% of tokens (313 patches) with the highest contribution to the 'Bowling' class.

As shown in Tables 1 and 2, this aggressive ablation had a negligible effect on the model's top class predictions. For the 'Strike' video (Table 1), the model retained the 'Bowling' classification with a negligible logit decrease (-0.34). Similarly, for the 'Gutter' video (Table 2), the confidence reduced slightly (-0.02) indicating no real change in model's perception post ablations.

This creates a paradox: the model calculates a strong outcome signal (Section 4.2), yet the classification circuit is highly distributed and does not strictly require the salient action features as seen after component ablations. This resilience provides strong causal evidence that the outcome circuit is a "hidden" mechanism that operates independently of the explicit classification task, motivating the use of Activation Patching (Section 4.4) to precisely identify the components responsible for this hidden computation.

\subsection{4.4. The Causal Mechanism: A Division of Labor}
\label{sec:headings}

Having localized the internal outcome signal and confirmed its functional independence from the explicit classification task (via ablation), activation patching was employed to reverse-engineer the model's construction mechanism. We systematically patch individual components (Attention vs. MLP blocks) from the 'strike' run into the 'gutter' run, quantifying the causal contribution of each block by measuring the percentage of the final signal recovered.

Leveraging the 'Success vs. Failure' signal delta ( $\Delta_{\text{orig}}$) as defined in the first equation: 
$\Delta_{\text{orig}} = act_{\text{strike}} - act_{\text{gutter}}$. Then a new delta is calculated (the patched delta, $\Delta_{\text{patch}}$) by running the 'gutter' video with a patched component and subtracting the original 'gutter' run activations: $\Delta_{\text{patch}} = act_{\text{patched\_gutter}} - act_{\text{gutter}}$. The percentage of Signal Recovery is then calculated by comparing these two deltas:
Formalizing 'Signal Recovery' as:

\begin{equation}
\label{eq:signal_recovery}
\text{Recovery} (\%) =
\frac{\|\Delta_{\text{patch}}\|}{\|\Delta_{\text{strike}}\|}
\cdot \text{sign}(\Delta_{\text{patch}} \cdot \Delta_{\text{strike}})
\times 100
\end{equation}

where $\Delta(act)$ is the L2 norm of the 'strike' minus 'gutter' activation at Layer 11, and the sign correction accounts for cases where the signal is inverted.

The results as seen in Figure 7 reveal a fundamental division of labor:

\subsubsection{Attention Heads act as Evidence Gatherers:} Patching attention blocks makes a significant but partial contribution (37-54\% signal recovery). Their role is to move relevant spatio-temporal evidence onto the residual stream.

\subsubsection{MLPs act as Concept Composers:} Patching single MLP blocks consistently recovers a larger proportion of the signal (42–60\%) from layers 4 through 9, identifying them as the primary drivers of the outcome representation. The fact that no single component achieves 100\% recovery confirms that the circuit is distributed: the model constructs the 'Success' outcome cumulatively across layers. This distributed nature explains the robustness observed in the earlier ablation experiments since the signal is not bottlenecked in a single block, the model remains resilient to individual component failures.

\begin{table}[H]
\centering
\begin{tabularx}{0.98\linewidth}{|c|c|c|c|}
\hline
\textbf{Layer} & \textbf{Component} & \textbf{Label} & \textbf{\% Recovery} \\
\hline
4  & Attention & L4 Attention  & 54.41 \\
4  & MLP       & L4 MLP        & 60.17 \\
5  & Attention & L5 Attention  & 50.22 \\
5  & MLP       & L5 MLP        & 57.49 \\
6  & Attention & L6 Attention  & 43.62 \\
6  & MLP       & L6 MLP        & 49.11 \\
7  & Attention & L7 Attention  & 40.38 \\
7  & MLP       & L7 MLP        & 42.55 \\
8  & Attention & L8 Attention  & 37.72 \\
8  & MLP       & L8 MLP        & 42.10 \\
9  & Attention & L9 Attention  & 44.43 \\
9  & MLP       & L9 MLP        & 58.66 \\
10 & Attention & L10 Attention & 47.61 \\
10 & MLP       & L10 MLP       & 43.39 \\
\hline
\end{tabularx}

\caption{Patching Experiment Results showing causal effects per layer and component.}
\label{tab:patching-results}
\end{table}

\begin{figure*}[t]
  \centering
  \includegraphics[width=0.83\textwidth]{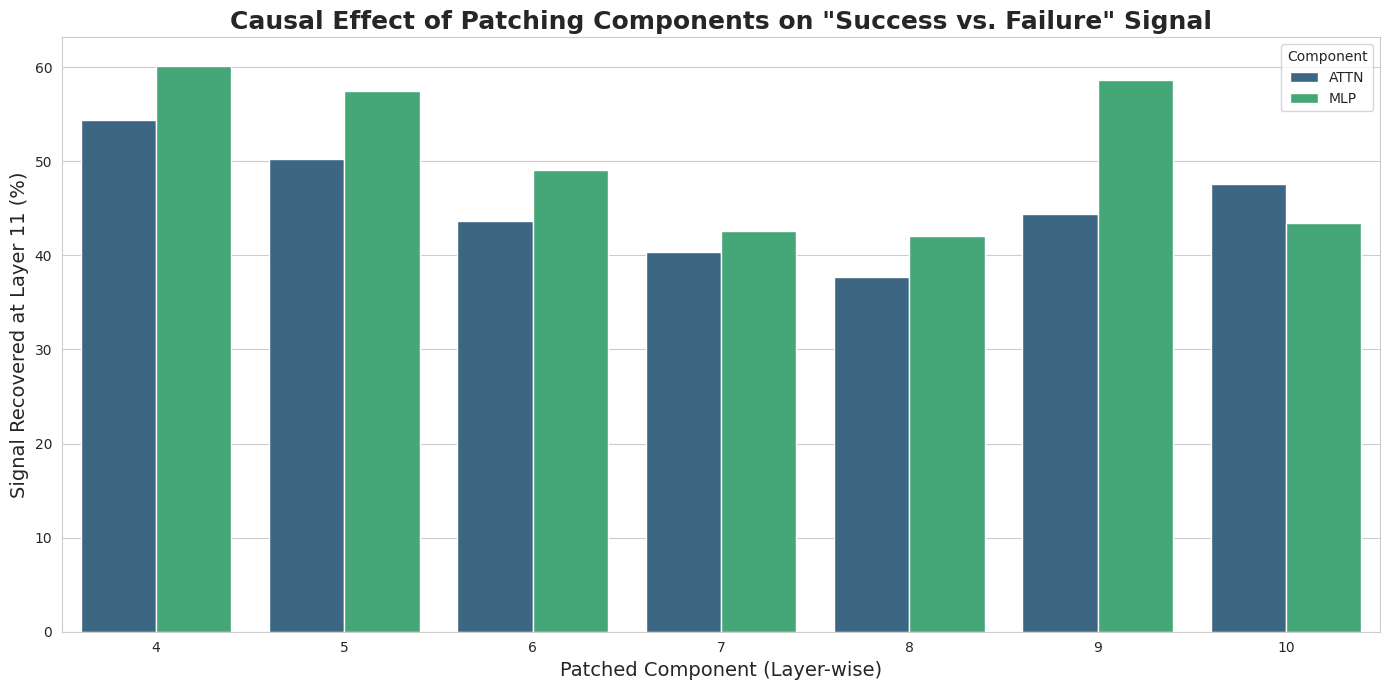}
  \caption{Causal effect of patching individual components on the layer 11 signal. MLP blocks are causally sufficient to create the outcome signal.}
  \label{fig:fig2}
\end{figure*}

\section{5. Discussion \& Limitations}
\label{sec:others}

\subsection{5.1. Conclusion}
This investigation successfully reverse-engineered a core computational pattern in VideoViT. The model represents nuanced action outcomes not via a single, fragile circuit, but through a robust, distributed algorithm where Attention Gathers, MLPs Compose. This discovery causally explains the model's invariance to simple ablation experiments; the resilience comes from the powerful, redundant cascade of its MLP blocks.  The results highlight how internal outcome representations can persist invisibly, motivating mechanistic oversight tools for safe model deployment.

\subsection{5.2. Limitations}
Since the experiments are foundational, a primary limitation is the scope. The findings are based on a single minimal pair of videos ("strike" vs. "gutter") and one specific pre-trained model architecture. Given the high dimensionality of videos (spatio-temporal tokens), a 'deep-dive' case study approach is adopted. This allowed for a granular, component-level causal analysis that involved patching of specific Attention heads and MLP blocks, which is computationally prohibitive to perform manually at scale.

This controlled design implies that while the internal validity of the circuit for this specific pair is established, we have not yet quantified the circuit's invariance to scene-specific details. We cannot strictly rule out the possibility that the identified circuit relies on features specific to this pair (e.g., background textures) rather than fully on a generalized semantic concept. However, while not generalizable across samples yet, the existence proof of "hidden cognition" is sufficient to question the safety of deployed AI systems, as it confirms that models can harbor internal representations that are effectively invisible to standard output monitoring and that are independent of original training tasks.

\subsection{5.3. Future Directions}

The most critical next step is to validate the "Attention Gathers, MLPs Compose" hypothesis at scale. Future work will move beyond manual patching to Automated Circuit Discovery (ACDC) techniques, allowing the extraction of this circuit across a statistically significant sample of the Kinetics-400 dataset. This will help to distinguish whether the "compositional MLP" pattern is a universal mechanism for processing outcomes, specific to trajectory-based actions like bowling, or merely an artifact of specific sample pairs. Additionally, an objective is to benchmark these mechanistic findings against standard interpretability baselines, such as Integrated Gradients, to empirically demonstrate that mechanistic techniques offer superior fidelity in isolating semantic circuits within the video domain compared to gradient-based attribution methods. Finally, investigating similar compositional mechanism in different architectures (e.g., TimeSformer) will shed light on how model architecture and patch size influence the distribution of labor between attention and MLPs.

\section{6. Broader Impact Statement}
This work serves as an empirical case study to detect "hidden cognition" in AI models, where the internal representations of a model (e.g., "success" vs. "failure") are more nuanced than its final output, which is a simple human-action class of "bowling". The techniques used offer a path towards scalable oversight to identify discrepancies between a model’s behavior and its internal state, a key challenge in AI safety and trustworthy AI. The finding that the outcome-concept is computed robustly and redundantly by an MLP cascade suggests that simple safety interventions, such as removing a single "harmful" component, are likely to fail. This highlights the need for more sophisticated approaches before such AI systems are deemed safe and deployable.

\bibliography{references}  

@misc{arnab2021vivit,
  author    = {Arnab,Anurag and Dehghani, Mostafa and  Heigold, Georg and  Sun, Chen and  Luci, Mario and  Schmid, Cordelia},
  title     = "{ViViT: A Video Vision Transformer}",
  booktitle = "{Proceedings of the IEEE/CVF International Conference on Computer Vision (ICCV)}",
  year      = {2021},
  pages     = "6816--6826",
  doi       = {10.1109/ICCV48922.2021.00676}
}

@misc{vonEschenbach2021transparency,
  author  = {von Eschenbach, Warren J.},
  title   = {Transparency and the Black Box Problem: Why We Do Not Trust AI},
  journal = {Philosophy \& Technology},
  volume  = {34},
  number  = {4},
  pages   = {1607--1622},
  year    = {2021},
  doi     = {10.1007/s13347-021-00477-0}
}

@misc{bereska2024mechanistic,
  title        = {Mechanistic Interpretability for AI Safety -- A Review},
  author       = {Bereska, Leonard and Gavves, Efstratios},
  year         = {2024},
  eprint       = {2404.14082},
  archivePrefix= {arXiv},
  primaryClass = {cs.AI}
}

@misc{olah2020zoom,
  title   = {Zoom In: An Introduction to Circuits},
  author  = {Olah, Chris and Cammarata, Nick and Schubert, Ludwig and Goh, Gabriel and Petrov, Michael and Carter, Shan},
  journal = {Distill},
  year    = {2020},
  doi     = {10.23915/distill.00024},
  url     = {https://distill.pub/2020/circuits/zoom-in}
}

@misc{anichenko2024interpretable,
  title={Interpretable Action Recognition on Hard to Classify Actions},
  author={Anichenko, Anastasia and Guerin, Frank and Gilbert, Andrew},
  journal={arXiv preprint arXiv:2409.13091},
  year={2024}
}

@misc{kay2017kinetics,
  title={The kinetics human action video dataset},
  author={Kay, Will and Carreira, Joao and Simonyan, Karen and Zhang, Brian and Hillier, Chloe and Vijayanarasimhan, Sudheendra and Viola, Fabio and Green, Tim and Back, Trevor and Natsev, Paul and others},
  journal={arXiv preprint arXiv:1705.06950},
  year={2017}
}

@misc{vaswani2017attention,
  title     = {Attention Is All You Need},
  author    = {Vaswani, Ashish and Shazeer, Noam and Parmar, Niki and Uszkoreit, Jakob and Jones, Llion and Gomez, Aidan N. and Kaiser, {\L}ukasz and Polosukhin, Illia},
  booktitle = {Advances in Neural Information Processing Systems (NeurIPS)},
  year      = {2017},
  pages     = {5998--6008}
}

@misc{dosovitskiy2020image,
  title     = {An Image is Worth 16x16 Words: Transformers for Image Recognition at Scale},
  author    = {Dosovitskiy, Alexey and Beyer, Lucas and Kolesnikov, Alexander and Weissenborn, Dirk and Zhai, Xiaohua and Unterthiner, Thomas and Dehghani, Mostafa and Minderer, Matthias and Heigold, Georg and Gelly, Sylvain and Uszkoreit, Jakob and Houlsby, Neil},
  booktitle = {International Conference on Learning Representations (ICLR)},
  year      = {2021},
  eprint    = {2010.11929},
  archivePrefix = {arXiv},
  primaryClass  = {cs.CV}
}

@misc{meyers2023safety,
  title={Safety-critical computer vision: an empirical survey of adversarial evasion attacks and defenses on computer vision systems},
  author={Meyers, Charles and L{\"o}fstedt, Tommy and Elmroth, Erik},
  journal={Artificial Intelligence Review},
  volume={56},
  number={Suppl 1},
  pages={217--251},
  year={2023},
  publisher={Springer}
}

@misc{gabriel2020artificial,
  title={Artificial intelligence, values, and alignment},
  author={Gabriel, Iason},
  journal={Minds & Machines 30, 411–437 (2020)},
  volume={30},
  number={3},
  pages={411--437},
  year={2020},
  publisher={Springer}
}

@misc{burns2022discovering,
  title={Discovering latent knowledge in language models without supervision},
  author={Burns, Collin and Ye, Haotian and Klein, Dan and Steinhardt, Jacob},
  journal={arXiv preprint arXiv:2212.03827},
  year={2022}
}

@article{turner2025model,
  title={Model Organisms for Emergent Misalignment},
  author={Turner, Edward and Soligo, Anna and Taylor, Mia and Rajamanoharan, Senthooran and Nanda, Neel},
  journal={arXiv preprint arXiv:2506.11613},
  year={2025}
}

@misc{mallen2023eliciting,
  title        = {Eliciting Latent Knowledge from Quirky Language Models},
  author       = {Mallen, Alex and Brumley, Madeline and Kharchenko, Julia and Belrose, Nora},
  year         = {2023},
  eprint       = {2312.01037},
  archivePrefix= {arXiv},
  primaryClass = {cs.LG}
}

@misc{cywinski2025eliciting,
  title        = {Eliciting Secret Knowledge from Language Models},
  author       = {Cywiński, Bartosz and Ryd, Emil and Wang, Rowan and Rajamanoharan, Senthooran and Nanda, Neel and Conmy, Arthur and Marks, Samuel},
  year         = {2025},
  eprint       = {2510.01070},
  archivePrefix= {arXiv},
  primaryClass = {cs.CL}
}

@article{sammani2023visualizing,
  title={Visualizing and understanding contrastive learning},
  author={Sammani, Fawaz and Joukovsky, Boris and Deligiannis, Nikos},
  journal={IEEE Transactions on Image Processing},
  volume={33},
  pages={541--555},
  year={2023},
  publisher={IEEE}
}

@article{joseph2023vit,
  title={Vit prisma: A mechanistic interpretability library for vision transformers},
  author={Joseph, Sonia},
  journal={GitHub repository},
  pages={30},
  year={2023}
}

@article{zhao2024explainability,
  title={Explainability for large language models: A survey},
  author={Zhao, Haiyan and Chen, Hanjie and Yang, Fan and Liu, Ninghao and Deng, Huiqi and Cai, Hengyi and Wang, Shuaiqiang and Yin, Dawei and Du, Mengnan},
  journal={ACM Transactions on Intelligent Systems and Technology},
  volume={15},
  number={2},
  pages={1--38},
  year={2024},
  publisher={ACM New York, NY}
}

@article{heimersheim2404use,
  title={How to use and interpret activation patching},
  author={Heimersheim, Stefan and Nanda, Neel},
  journal={arXiv preprint arXiv:2404.15255},
  year={2024},
  eprint       = {2404.15255},
  archivePrefix= {arXiv},
  primaryClass = {cs.LG}
}

@article{meloux2025everything,
  title={Everything, Everywhere, All at Once: Is Mechanistic Interpretability Identifiable?},
  author={M{\'e}loux, Maxime and Maniu, Silviu and Portet, Fran{\c{c}}ois and Peyrard, Maxime},
  journal={arXiv preprint arXiv:2502.20914},
  year={2025}
}

\end{document}